\documentclass{article}

 \usepackage[dblblindworkshopeurips, final]{neurips_2025}

\usepackage[utf8]{inputenc} 
\usepackage[T1]{fontenc}    
\usepackage{hyperref}       
\usepackage{url}            
\usepackage{booktabs}       
\usepackage{amsfonts}       
\usepackage{nicefrac}       
\usepackage{microtype}      
\usepackage{xcolor}         

\usepackage{amsmath,amssymb}
\usepackage{algorithm}
\usepackage{algpseudocode}

\title{Amortized Causal Discovery with Prior-Fitted Networks}

\author{
Mateusz Sypniewski\thanks{Equal contribution.}\\
  University of Cambridge\\
  \And
  Mateusz Olko$^{*}$\\
  University of Warsaw\\
  IDEAS NCBR\\
  \texttt{mateusz.olko@gmail.com} \\
  \AND
Mateusz Gajewski\\
  Poznan University of Technology\\
  IDEAS NCBR\\
  \texttt{mg96272@gmail.com} \\
  \And
  Piotr Miłoś\\
University of Warsaw\\
}

\begin{document}

\maketitle

\begin{abstract}
In recent years, differentiable penalized likelihood methods have gained popularity, optimizing the causal structure by maximizing its likelihood with respect to the data. However, recent research has shown that errors in likelihood estimation, even on relatively large sample sizes, disallow the discovery of proper structures. 
We propose a new approach to amortized causal discovery that addresses the limitations of likelihood estimator accuracy. Our method leverages Prior-Fitted Networks (PFNs) to amortize data-dependent likelihood estimation, yielding more reliable scores for structure learning. Experiments on synthetic, simulated  and real-world datasets show significant gains in structure recovery compared to standard baselines. Furthermore, we demonstrate directly that PFNs provide more accurate likelihood estimates than conventional neural network–based approaches.
\end{abstract}

\section{Introduction}

Causal discovery is at the heart of every scientific inquiry. Recent years brought significant advancements in causal discovery methods' scalability and applicability. 

One of the most prominent modern approaches to causal discovery is the class of so called differentiable penalized likelihood methods \citep{dcdi, notears}.
 They define a causal discovery problem as constrained optimization of the score based on the likelihood of the data. Therefore, they consist of two components: likelihood estimation and structure optimization.

However, recent research has identified critical limitations in this paradigm. \citet{Olko2025SinceFaithfulness} have shown that errors in likelihood estimation, even on relatively large sample sizes, disallow discovery of proper structures. The margin for error quickly decreases with the increasing graph size and density leading to unrealistic data volume requirements, which significantly constrains the applicability of current approaches in real-world scenarios.

In this work we propose a novel approach to amortized causal discovery that addresses the limitation of estimator accuracy. Our approach uses pre-trained networks to amortize the data-dependent  likelihood estimation stage and increase its accuracy.  The improved estimators are used in downstream causal structure optimization using policy gradients. Additionally, the approach allows to decouple training of likelihood estimators and graph structure optimization, leading to more stable training. 

To enable this, we leverage Prior-Data Fitted Networks (PFNs), which use in-context learning and large-scale transformer architectures to approximate a wide range of posterior distributions \citep{Muller2022PFN}. PFNs gained attention after the release of TabPFN, which achieved state-of-the-art performance on tabular tasks, outperforming traditional methods such as gradient-boosted trees \citep{Hollmann2023TabPFN, Hollmand2025TabPFN2}.
These models allow for Bayesian inference at test time without retraining, making them ideal for our application.

 Our contributions include:
 \begin{itemize}
     \item We define a novel causal discovery method that leverage prior-fitted networks for straightforward and accurate likelihood estimates.
     \item Our method outperforms standard baselines in terms of accuracy on synthetic benchmark and matches their performance on real-world and simulated data.
     \item We provide qualitative comparison of the quality of likelihood estimates between PFN approach and standard NN based approach.
 \end{itemize}

\section{Preliminaries} 
\paragraph{Causal discovery \& penalized likelihood score}
The goal of causal discovery is to recover the underlying graph structure of the data-generating process. A common approach is to search for the graph that maximizes a penalized likelihood score:
\begin{equation} \label{eq:pen_likelihood}
    s(G, D) = \log p(D \mid G) - \lambda |G| ,
\end{equation}
where $G$ denotes the candidate graph, $D$ the observed dataset, $|G|$ the number of edges in $G$, and $\lambda > 0$ a regularization coefficient that penalizes overly dense structures. 

Without further assumptions, the true data-generating graph is not uniquely identifiable; one can only recover its Markov equivalence class (MEC), i.e., the set of graphs that entail the same conditional independence relations. Nevertheless, optimizing the score in Equation~\ref{eq:pen_likelihood} with sufficiently small $\lambda$ is guaranteed to recover a member of the MEC of the true structure \citep{dcdi}.

\paragraph{TabPFN}
TabPFN is a pre-trained Transformer for supervised classification and regresssion on tabular data \citep{Hollmann2023TabPFN}. TabPFN is trained once on synthetically generated tasks and can subsequently be applied to unseen datasets via in-context learning in a single forward pass. TabPFN builds on Prior-Data Fitted Networks \citep{Muller2022PFN}, during inference it receives both the training samples and the test inputs and directly outputs an approximation of the posterior predictive distribution, effectively performing Bayesian inference in one step.

The PFN training objective is defined on synthetic datasets sampled from a prior \(p(D)\). For a single test point
\(\{(x_{\text{test}}, y_{\text{test}})\} = D_{\text{test}}\),
the loss minimized during training is
\[
\mathcal{L}_{\text{PFN}} =
\mathbb{E}_{\{(x_{\text{test}}, y_{\text{test}})\} \cup D_{\text{train}} \sim p(D)}
\left[ - \log q_{\theta}(y_{\text{test}} \mid x_{\text{test}}, D_{\text{train}}) \right],
\]
where $\theta$ are parameters of the PFN model.

\section{Method} 

The method is outlined in Algorithm~\ref{alg:method}. For ease of explanation,  we split the description into three components.

\paragraph{Parametrization of the graph posterior}
We use flexible posterior definition of \citet{Charpentier2022Differentiable}. The posterior is parametrized by two continuous matrices $P, W \in R^{d\times d}$. Through the gumbel softmax  and softsort procedures they are used to sample binary edges $E$ and permutation $\Pi$ matrices. Those are then combined with an upper triangular matrix $M$ to provide the adjacency matrix of sampled DAG $A = E \cdot (\Pi^T \times M \times \Pi)$. This representation method generates acyclic graphs by design while maintaining differentiability. Additionally it is extremely expressive and can represent vide range of stochastic posteriors over DAGs.

\paragraph{Likelihood score estimation}
To estimate the likelihood of a specific graph $G$ on dataset $D$ the data is first split into train and estimation sets \footnote{The split is an arbitrary decision, it is also possible to use cross validation to better leverage the available data.}, $D_{train}$ and  $D_{est}$ accordingly.
Than for each variable $i$ a set of parents $\text{pa}_G(i)$ is extracted and a PFN model is fitted in context, only using columns $\text{pa}_G(i)$, to obtain the likelihood estimates $\log p_i(D_{est} | D_{train}, \text{pa}_G(i))$. The score is computed as the sum over all variables in the graph:
\begin{equation}
    \hat{s}(G, D) = \sum_{i \in V} \log p_i(D_{est} | D_{train}, \text{pa}_G(i)) - \lambda |G|
\end{equation}

\paragraph{Optimization}
Since the likelihood estimates produced by PFN model are non-differentiable, while the graph-sampling procedure is differentiable, we can naturally cast the optimization problem as a reinforcement learning task solvable with policy gradient methods. In this formulation, the DAG sampling mechanism defines the policy, the action space consists of all possible DAGs, and the PFN likelihood estimate serves as the reward. Each episode reduces to a single decision step.  

In our experiments, we use Proximal Policy Optimization (PPO) \citep{schulman_proximal_2017}, though any policy gradient method would work. PPO has achieved remarkable success in recent years thanks to both its strong performance and ease of implementation. It introduces constraints on how much the updated policy is allowed to deviate from the previous one after each step. 
The PPO clipped objective is given by

\[
L^{\text{CLIP}}(\theta) = \mathbb{E} \Big[ \min \big( r(\theta)\, A, \; \text{clip}(r(\theta), 1 - \epsilon, 1 + \epsilon)\, A \big) \Big], \; \text{where} \;  r(\theta) = \frac{\pi_\theta(a)}{\pi_{\theta_{\text{old}}}(a)}.
\]

The $r(\theta)$ is the probability ratio between the new and old policies. $A$ is the advantage function, which we define as $
A = R - b$
where $R$ is the current reward and $b$ is the exponential moving average of historical rewards.  

\begin{algorithm}[t]
\caption{Graph Posterior Optimization with PPO}
\label{alg:method}
\begin{algorithmic}[1]
\Require Dataset $D$, regularization parameter $\lambda$, number of iterations $T$, PPO update frequency $K$
\State Initialize posterior parameters $(P, W)$
\State Split $D$ into $D_{\text{train}}$ and $D_{\text{est}}$
\For{$t = 1, \dots, T$}
    \State \textbf{Data Acquisition:}
    \For{$k = 1, \dots, K$}
        \State Sample adjacency matrix $A_{t,k}$ from posterior $(P, W)$ using Gumbel--Softmax and SoftSort
        \State Compute log-likelihood of sampled adjacency: $\log \pi_{(P,W)}(A_{t,k})$
        \For{each variable $i \in V$}
            \State Fit PFN model on $(D_{\text{train}}, \text{pa}_{A_{t,k}}(i))$
            \State Compute likelihood $\log p_i(D_{\text{est}} \mid D_{\text{train}}, \text{pa}_{A_{t,k}}(i))$
        \EndFor
        \State Compute reward (score): 
        \[
           \hat{s}(A_{t,k}, D) = \sum_{i \in V} \log p_i(D_{\text{est}} \mid D_{\text{train}}, \text{pa}_{A_{t,k}}(i)) - \lambda |A_{t,k}|
        \]
        \State Store $(A_{t,k}, \log \pi_{(P,W)}(A_{t,k}), \hat{s}(A_{t,k}, D))$ in trajectory buffer
    \EndFor
    \State \textbf{Policy Update:}
    \State Use PPO to update posterior parameters $(P, W)$ based on stored tuples $(A_{t,k}, \log \pi_{(P,W)}(A_{t,k}), \hat{s}(A_{t,k}, D))$
\EndFor
\State \Return Learned posterior over graphs
\end{algorithmic}
\end{algorithm}
\section{Experiments} 

\subsection{Evaluation of likelihood estimators} 
We begin with validation that amortized prediction with PFN provides more accurate likelihood estimates than MLPs trained specifically for this task. To this end we use data bootstraping to asses which estimation method gives more consistent results and compare the accuracy of score estimations using groud truth graph.

Specifically, for a fixed dataset size, we sample 30 training datasets and a heldout set of 10 000 samples from an SCM with Erdos-Renyi graph prior with 5 nodes and 5 edges and the functional relations initalized using random neural networks. Than, for each variable $i  \in V$ and each possible parent set $S \subseteq V\setminus\{i\}$, we fit an estimator using training dataset and compute average negative log likelihood of the samples from the held-out set. Using predictions across 30 datasets we compute bootstrap variance, and than aggregate it across all varaibles and parentsets using mean and median, see columns 1 and 2 in Table~\ref{tab:quality}. Additionally, we report average NLL value in column 3.

To evalute the accuracy of the estimation we compare estimated scores of all graphs with 5 nodes. For each dataset we compute how many structures obtained better score than the ground truth structure\footnote{As ground truth we select the structure from the MEC of the data generating strcture that obtained the best (the highest) score estimate.}. In column 4, we report this number averaged over 30 boostraped datasets.

\paragraph{Results} Across all dataset sizes, PFN-based estimation produces more consistent results, with substantially lower variance, than the MLP baseline, and it outperforms the MLP by a wide margin. Moreover, PFN estimates are markedly more accurate, enabling recovery of the ground-truth graph for all dataset sizes except the smallest (125 samples). Finally, the PFN yields lower estimated NLL values, suggesting that it produces a more concentrated posterior distribution.
These findings indicate that incorporating an appropriate prior can substantially enhance estimation quality. 

\begin{table}[tbp]
    \centering
    \begin{tabular}{l|cc|cc|cc|cc}
    \toprule
     & \multicolumn{2}{c}{Mean BV} & \multicolumn{2}{c}{Median BV} & \multicolumn{2}{c}{NLL value} & \multicolumn{2}{c}{\#Incorrect structures} \\
    \#Samples & MLP & PFN & MLP & PFN & MLP & PFN & MLP & PFN \\
    \midrule
    125  &  22.66 & 2.4e-03 & 0.60 & 2.1e-03 & 2.24 & -3.4e-01 & 3137.43 & 6.7e-02 \\
    250  &  0.62 & 7.8e-04 & 3.5e-02 & 5.1e-04 & 0.84 & -4.6e-01 & 55.03 & 0 \\
    500  &  8.9e-02 & 2.6e-04 & 1.5e-03 & 1.6e-04 & 0.51 & -5.5e-01 & 6.40 & 0 \\
    1000  &  2.4e-02 & 1.3e-04 & 4.4e-04 & 5.8e-05 & 0.38 & -6.0e-01 & 2.27 & 0 \\
    2000  &  1.6e-02 & 3.6e-04 & 1.8e-04 & 2.1e-05 & 0.29 & -6.4e-01 & 1.76 & 0 \\
    \bottomrule
    \end{tabular}
    \caption{Results of the estimator quality evaluation.}
    \label{tab:quality}
\end{table}

\subsection{Quantitative evaluations}
\paragraph{Datasets.}
We evaluate our approach on three categories of datasets. First, we consider synthetic settings where graphs are sampled from the Erdős--Rényi class, functional relations are modeled using randomly initialized neural networks, and additive Gaussian noise with different variances is assumed, following standard practice in the literature \citep{dcdi, dibs, sdcd, bayesdag}. We consider a range of graph sizes and densities, as detailed in Table~\ref{tab:benchmark_results}. 

Second, we evaluate on datasets generated by the SERGIO simulator, which models realistic biological relations~\citep{dibaeinia2020sergio}, with underlying graph structures sampled from a scale-free distribution and functional relationships generated using expert-designed differentiable equations.

Third, we use a dataset generated using Causal Chambers~\citep{gamella2025causal}—a real-world physical system where a ground-truth DAG is known—consisting of one 20-node graph.

\paragraph{Baselines.}

We compare the proposed method to two causal discovery methods: PC~\citep{spirtes2000causation} and DCDI~\citep{dcdi}. PC (Peter-Clark) performs conditional independence tests to recover the true graph; we use the FisherZ conditional independence test. DCDI (Differentiable Causal Discovery from Interventional Data) is a continuous neural method that leverages expressive neural architectures for causal discovery.

\paragraph{Metrics.}
We report performance using Structural Hamming Distance (SHD), since causal discovery in our benchmark is not identifiable beyond the Markov equivalence class, we evaluate distances between the completed partially directed acyclic graphs (CPDAGs) of the estimated and ground-truth structures. 

\begin{table}[tbp]
    \centering
    \begin{tabular}{c|c|c|c|c}
    \toprule
    Method & ER(10, 20) & ER(30, 60) & Chamber p20 & Sergio p10 \\
    \midrule

PC Gaussian &	 18.40 \tiny(15.20, 20.60)&  59.80 \tiny(54.20, 64.50) & 38 &30.90 \tiny(28.20, 32.70)\\
DCDI 	& 17.30 \tiny(13.90, 21.40) & 38.90 \tiny(29.90, 47.30)  & 49 & 29.10 \tiny(25.80, 31.80)\\
ACD 	 &1.44 \tiny(0.87, 2.99) & 17.52 \tiny(13.53, 21.84) & 44 & 31.56 \tiny(29.33, 33.33)\\
\bottomrule
    \end{tabular}
    \caption{Benchmark results on synthetic and SERGIO datasets. Values in brackets describe 95\% bootstrap confidence intervals.}    \label{tab:benchmark_results}
\end{table}

\paragraph{Results.}
Table~\ref{tab:benchmark_results} summarizes the benchmark results. 
Our method (ACD) substantially outperforms other methods on synthetic datasets, achieving the lowest structural Hamming distance on both ER graphs.
On the Causal Chambers dataset, ACD performs better than DCDI but slightly worse than PC. 
On the semi-synthetic SERGIO dataset, ACD performs comparably to DCDI and PC, with overlapping confidence intervals.
These results highlight the benefit of incorporating prior-informed likelihood estimation into the causal discovery pipeline.

\paragraph{Prior Alignment.}
We observe drastic improvements on synthetic datasets, which we hypothesize stem from TabPFN's training distribution. Specifically, TabPFN was pretrained on synthetic Bayesian networks and structural causal models~\citep{Hollmand2025TabPFN2}, resulting in a prior that is well-aligned with the functional relationships in our synthetic benchmarks. In contrast, performance on Causal Chambers and SERGIO datasets is comparatively weaker, suggesting a misalignment between TabPFN's prior and the relationships in these real-world and semi-synthetic systems. This observation points to an exciting direction for future work: fine-tuning existing priors or constructing new priors better aligned with specific domains to improve causal discovery performance across diverse data-generating processes.
\section{Discussion} 

Since our method amortizes only the likelihood prediction component of the causal discovery pipeline, while relying on structure optimization over a well-established score, it retains the identification guarantees established by \citet{dcdi}. This represents a significant advantage over recently proposed fully amortized approaches and methods based on large language models, whose theoretical properties remain poorly understood. At the same time, our approach is broadly applicable: it can be deployed to a wide range of problems with comparable ease and requires only minimal hyperparameter tuning.

\bibliography{preludium}

@article{Olko2025SinceFaithfulness,
  author       = {Mateusz Olko and
                  Mateusz Gajewski and
                  Joanna Wojciechowska and
                  Mikolaj Morzy and
                  Piotr Sankowski and
                  Piotr Milos},
  title        = {Since Faithfulness Fails: The Performance Limits of Neural Causal
                  Discovery},
  journal      = {CoRR},
  volume       = {abs/2502.16056},
  year         = {2025},
  url          = {https://doi.org/10.48550/arXiv.2502.16056},
  doi          = {10.48550/ARXIV.2502.16056},
  eprinttype    = {arXiv},
  eprint       = {2502.16056},
  bibsource    = {dblp computer science bibliography, https://dblp.org}
}

@inproceedings{Muller2022PFN,
  author       = {Samuel M{\"{u}}ller and
                  Noah Hollmann and
                  Sebastian Pineda{-}Arango and
                  Josif Grabocka and
                  Frank Hutter},
  title        = {Transformers Can Do Bayesian Inference},
  booktitle    = {The Tenth International Conference on Learning Representations, {ICLR}
                  2022, Virtual Event, April 25-29, 2022},
  publisher    = {OpenReview.net},
  year         = {2022},
  url          = {https://openreview.net/forum?id=KSugKcbNf9},
  bibsource    = {dblp computer science bibliography, https://dblp.org}
}

@inproceedings{Charpentier2022Differentiable,
  author       = {Bertrand Charpentier and
                  Simon Kibler and
                  Stephan G{\"{u}}nnemann},
  title        = {Differentiable {DAG} Sampling},
  booktitle    = {The Tenth International Conference on Learning Representations, {ICLR}
                  2022, Virtual Event, April 25-29, 2022},
  publisher    = {OpenReview.net},
  year         = {2022},
  url          = {https://openreview.net/forum?id=9wOQOgNe-w},
  bibsource    = {dblp computer science bibliography, https://dblp.org}
}

@article{Hollmand2025TabPFN2,
  author       = {Noah Hollmann and
                  Samuel M{\"{u}}ller and
                  Lennart Purucker and
                  Arjun Krishnakumar and
                  Max K{\"{o}}rfer and
                  Shi Bin Hoo and
                  Robin Tibor Schirrmeister and
                  Frank Hutter},
  title        = {Accurate predictions on small data with a tabular foundation model},
  journal      = {Nat.},
  volume       = {637},
  number       = {8044},
  pages        = {319--326},
  year         = {2025},
  url          = {https://doi.org/10.1038/s41586-024-08328-6},
  doi          = {10.1038/S41586-024-08328-6},
  bibsource    = {dblp computer science bibliography, https://dblp.org}
}

@inproceedings{Hollmann2023TabPFN,
  author       = {Noah Hollmann and
                  Samuel M{\"{u}}ller and
                  Katharina Eggensperger and
                  Frank Hutter},
  title        = {TabPFN: {A} Transformer That Solves Small Tabular Classification Problems
                  in a Second},
  booktitle    = {The Eleventh International Conference on Learning Representations,
                  {ICLR} 2023, Kigali, Rwanda, May 1-5, 2023},
  publisher    = {OpenReview.net},
  year         = {2023},
  url          = {https://openreview.net/forum?id=cp5PvcI6w8\_},
  bibsource    = {dblp computer science bibliography, https://dblp.org}
}

@inproceedings{dcdi,
  author       = {Philippe Brouillard and
                  S{\'{e}}bastien Lachapelle and
                  Alexandre Lacoste and
                  Simon Lacoste{-}Julien and
                  Alexandre Drouin},
  editor       = {Hugo Larochelle and
                  Marc'Aurelio Ranzato and
                  Raia Hadsell and
                  Maria{-}Florina Balcan and
                  Hsuan{-}Tien Lin},
  title        = {Differentiable Causal Discovery from Interventional Data},
  booktitle    = {Advances in Neural Information Processing Systems 33: Annual Conference
                  on Neural Information Processing Systems 2020, NeurIPS 2020, December
                  6-12, 2020, virtual},
  year         = {2020},
  url          = {https://proceedings.neurips.cc/paper/2020/hash/f8b7aa3a0d349d9562b424160ad18612-Abstract.html},
  bibsource    = {dblp computer science bibliography, https://dblp.org}
}

@inproceedings{notears,
  author       = {Xun Zheng and
                  Bryon Aragam and
                  Pradeep Ravikumar and
                  Eric P. Xing},
  editor       = {Samy Bengio and
                  Hanna M. Wallach and
                  Hugo Larochelle and
                  Kristen Grauman and
                  Nicol{\`{o}} Cesa{-}Bianchi and
                  Roman Garnett},
  title        = {DAGs with {NO} {TEARS:} Continuous Optimization for Structure Learning},
  booktitle    = {Advances in Neural Information Processing Systems 31: Annual Conference
                  on Neural Information Processing Systems 2018, NeurIPS 2018, December
                  3-8, 2018, Montr{\'{e}}al, Canada},
  pages        = {9492--9503},
  year         = {2018},
  url          = {https://proceedings.neurips.cc/paper/2018/hash/e347c51419ffb23ca3fd5050202f9c3d-Abstract.html},
}

@inproceedings{bayesdag,
  author       = {Yashas Annadani and
                  Nick Pawlowski and
                  Joel Jennings and
                  Stefan Bauer and
                  Cheng Zhang and
                  Wenbo Gong},
  editor       = {Alice Oh and
                  Tristan Naumann and
                  Amir Globerson and
                  Kate Saenko and
                  Moritz Hardt and
                  Sergey Levine},
  title        = {BayesDAG: Gradient-Based Posterior Inference for Causal Discovery},
  booktitle    = {Advances in Neural Information Processing Systems 36: Annual Conference
                  on Neural Information Processing Systems 2023, NeurIPS 2023, New Orleans,
                  LA, USA, December 10 - 16, 2023},
  year         = {2023},
  url          = {http://papers.nips.cc/paper\_files/paper/2023/hash/05cf28e3d3c9a179d789c55270fe6f72-Abstract-Conference.html},
  bibsource    = {dblp computer science bibliography, https://dblp.org}
}

@inproceedings{dibs,
  author       = {Lars Lorch and
                  Jonas Rothfuss and
                  Bernhard Sch{\"{o}}lkopf and
                  Andreas Krause},
  editor       = {Marc'Aurelio Ranzato and
                  Alina Beygelzimer and
                  Yann N. Dauphin and
                  Percy Liang and
                  Jennifer Wortman Vaughan},
  title        = {DiBS: Differentiable Bayesian Structure Learning},
  booktitle    = {Advances in Neural Information Processing Systems 34: Annual Conference
                  on Neural Information Processing Systems 2021, NeurIPS 2021, December
                  6-14, 2021, virtual},
  pages        = {24111--24123},
  year         = {2021},
  url          = {https://proceedings.neurips.cc/paper/2021/hash/ca6ab34959489659f8c3776aaf1f8efd-Abstract.html},
  bibsource    = {dblp computer science bibliography, https://dblp.org}
}

@inproceedings{sdcd,
  author       = {Achille Nazaret and
                  Justin Hong and
                  Elham Azizi and
                  David M. Blei},
  title        = {Stable Differentiable Causal Discovery},
  booktitle    = {Forty-first International Conference on Machine Learning, {ICML} 2024,
                  Vienna, Austria, July 21-27, 2024},
  publisher    = {OpenReview.net},
  year         = {2024},
  url          = {https://openreview.net/forum?id=JJZBZW28Gn},
  bibsource    = {dblp computer science bibliography, https://dblp.org}
}

@article{gamella2025causal,
  title={Causal chambers as a real-world physical testbed for AI methodology},
  author={Gamella, Juan L and Peters, Jonas and B{\"u}hlmann, Peter},
  journal={Nature Machine Intelligence},
  pages={1--12},
  year={2025},
  publisher={Nature Publishing Group UK London}
}

@misc{schulman_proximal_2017,
	title = {Proximal {Policy} {Optimization} {Algorithms}},
	url = {http://arxiv.org/abs/1707.06347},
	doi = {10.48550/arXiv.1707.06347},
	urldate = {2025-06-05},
	publisher = {arXiv},
	author = {Schulman, John and Wolski, Filip and Dhariwal, Prafulla and Radford, Alec and Klimov, Oleg},
	month = aug,
	year = {2017},
	note = {arXiv:1707.06347 [cs]},
}

@article{dibaeinia2020sergio,
  title={SERGIO: a single-cell expression simulator guided by gene regulatory networks},
  author={Dibaeinia, Payam and Sinha, Saurabh},
  journal={Cell systems},
  volume={11},
  number={3},
  pages={252--271},
  year={2020},
  publisher={Elsevier}
}

@book{spirtes2000causation,
  title={Causation, prediction, and search},
  author={Spirtes, Peter and Glymour, Clark N and Scheines, Richard},
  year={2000},
  publisher={MIT press}
}
\bibliographystyle{plainnat}

\end{document}